\documentclass{ecai} 

\usepackage{times}
\usepackage{soul}
\usepackage{url}
\usepackage[hidelinks]{hyperref}
\usepackage[utf8]{inputenc}
\usepackage[small]{caption}
\usepackage{booktabs}
\usepackage{algorithm}
\usepackage{algorithmic}
\usepackage[switch]{lineno}
\usepackage{latexsym}
\usepackage{amssymb}
\usepackage{amsmath}
\usepackage{amsthm}
\usepackage{booktabs}
\usepackage{enumitem}
\usepackage{graphicx}
\usepackage{color}
\usepackage[table]{xcolor}
\usepackage{lbargmacros}
\usepackage{amsfonts,amssymb}
\usepackage{xspace}
\usepackage{mathtools}
\usepackage{subcaption}
\usepackage{hyperref}
\usepackage{enumitem}
\usepackage{thm-restate}

\usepackage{tikz}
\usepackage{tikz-qtree}
\usetikzlibrary{patterns}
\usetikzlibrary{positioning}
\usetikzlibrary{arrows,automata,calc,backgrounds,fit}
\usetikzlibrary{shapes,snakes,bbox}
\usetikzlibrary{shapes.geometric}
\usepgflibrary{arrows}
\usepackage{tikz-qtree}
\usepackage{pifont}

\tikzset{
    >=stealth',
    args/.style={circle,draw=black,minimum size=0.7cm,inner sep=0,outer sep=0}    
}

\usetikzlibrary{fadings}

\newtheorem{example}{Example}

\newtheorem{corollary}{Corollary}

\theoremstyle{definition}
\newtheorem{definition}{Definition}
\newtheorem{principle}{Principle}

\newcommand{\BibTeX}{B\kern-.05em{\sc i\kern-.025em b}\kern-.08em\TeX}

\newcommand{\ie}{i.\,e.\xspace}				
\newcommand{\eg}{e.\,g.\xspace}				
\newcommand{\mt}[1]{\textcolor{red}{MT: #1}} 

\newcommand{\attacks}{R}				

\newcommand{\adm}{\textsf{adm}}
\newcommand{\co}{\textsf{co}}
\newcommand{\gr}{\textsf{gr}}
\newcommand{\pr}{\textsf{pr}}
\newcommand{\sst}{\textsf{sst}}
\newcommand{\id}{\textsf{id}}

\newcommand{\cf}{\cfree}
\newcommand{\na}{\naiv}

\newcommand{\ud}{\textsf{ud}}

\newcommand{\leftup[1]}{\prescript{#1}{}}

\begin{document}


\begin{frontmatter}


\paperid{1855} 

\title{Revisiting Vacuous Reduct Semantics for Abstract Argumentation (Extended Version)}


\author[A]{\fnms{Lydia}~\snm{Blümel}\thanks{Corresponding Author. Email: lydia.bluemel@fernuni-hagen.de.}}
\author[B]{\fnms{Matthias}~\snm{Thimm}\thanks{Corresponding Author. Email: matthias.thimm@fernuni-hagen.de.}}

\address[A,B]{Artificial Intelligence Group, University of Hagen, Germany}


\begin{abstract}
We consider the notion of a vacuous reduct semantics for abstract argumentation frameworks, which, given two abstract argumentation semantics $\sigma$ and $\tau$, refines  $\sigma$ (base condition) by accepting only those $\sigma$-extensions that have no non-empty $\tau$-extension in their reduct (vacuity condition). We give a systematic overview on vacuous reduct semantics resulting from combining  different admissibility-based and conflict-free semantics and present a principle-based analysis of vacuous reduct semantics in general. We provide criteria for the inheritance of principle satisfaction by a vacuous reduct semantics from its base and vacuity condition for established as well as recently introduced principles in the context of weak argumentation semantics. We also conduct a principle-based analysis for the special case of undisputed semantics.
\end{abstract}

\end{frontmatter}


\section{Introduction}
Abstract argumentation frameworks (AFs) \cite{Dung95} are a knowledge representation formalism that represents argumentation scenarios as directed graphs, where nodes represent arguments and edges represent conflicts between arguments. Works based on this basic AF representation are counted towards abstract argumentation, differentiating them from those with additional structural elements, \eg Weighted Argumentation Frameworks \cite{Coste-MarquisKMO12Weighted}, Constrained Argumentation Frameworks \cite{ConstrainedAF2006} or Structured Argumentation \cite{DungKT09ABA}.

There exist several different approaches regarding semantics for AFs. 
Here, we focus on the extension-based approach by which an argumentation semantics $\sigma$ is a mapping that assigns to an argumentation framework a set of extensions. Alternatives are, \eg labeling-based semantics \cite{ModgilC09label} and ranking-based semantics \cite{BonzonDKM16ranking}. A $\sigma$-extension is a set of arguments, which collectively satisfy the conditions specified by $\sigma$ \eg to have no attacks among themselves. \cite{ThimmAAAI23} proposes a method for combining two argumentation semantics $\sigma,\tau$ into a third we refer to as the $\tau$-vacuous reduct semantics to the base of $\sigma$, $\vac_\sigma(\tau)$. 
The extensions of $\vac_\sigma(\tau)$ satisfy two conditions: the base condition, requiring them to be $\sigma$-extensions, and the vacuity condition, demanding there exists no nonempty $\tau$-extension in their reduct \cite{BaumannBU2020}. 
Given a set of arguments $E$ of an AF $F$, the reduct is the restriction of $F$ to all arguments neither in nor attacked by $E$ and the attacks among them. 
The idea of vacuous reduct semantics is to check the reduct for relevant arguments, as specified by the vacuity condition $\tau$, and to only accept $\sigma$-extensions whose reduct proves to contain nothing of interest. For instance, using conflict-free semantics for the vacuity condition allows us to ignore the presence of self-attackers outside of our extension, a problem already mentioned by \cite{Dung95}, which has given rise to the class of weak argumentation semantics, \ie semantics violating admissibility in order to address this and other cases of unacceptable attackers \cite{BodanzaT09cogsem,BaumannBU2020,RienstraTW22nonadm}. In Section \ref{sec:sem} we extend the work on the undisputed semantics \cite{ThimmAAAI23} and the semantics discussed in \cite{BluemelThimmKR23}  by systematically investigating combinations of different admissibility- and conflict-free-based semantics into vacuous reduct semantics.

A prominent method for evaluating the different argumentation semantics is the principle-based analysis \cite{TorreVesic17Princ}. The term ``principle'' refers to a desirable property, although, depending on the use case intended for a semantics, the principles it should satisfy are subject to discussion. Apart from the established standard, a number of principles for weak argumentation semantics, which violate admissibility by design, have been proposed \cite{BaumannModularization020,BlumelUlbricht22DefeatOp}. In Section \ref{sec:princ} we derive general criteria for the satisfaction of a selection of principles from both categories by a vacuous reduct semantics.
 
Our contributions are as follows.
\begin{itemize}
\item We give a systematic overview on vacuous reduct semantics resulting from combining  different admissibility-based and conflict-free semantics (Section \ref{sec:sem})
\item We provide general criteria for principle satisfaction by a vacuous reduct semantics based on its base and vacuity condition (Section \ref{sec:princ})
\item We conduct a principle-based analysis for the undisputed semantics. (Section \ref{sec:princ})
\end{itemize}

Section \ref{sec:prelim} introduces the necessary terminology, Section \ref{sec:disc} discusses related work and Section \ref{sec:conc} concludes the paper.

\section{Preliminaries}
\label{sec:prelim}
We fix an infinite countable set $\arguniverse$ as the universe of arguments. We consider abstract argumentation frameworks by Dung [\citeyear{Dung95}], which can be represented as directed graphs.
\begin{definition}
    An \emph{Abstract Argumentation Framework (AF)} is a tuple $\argnetdef$, where $\allargs\subset\arguniverse$ is a finite set of arguments and $\allattacks\subseteq\allargs\times\allargs$ is the \emph{attack} relation. Let $\argnetuniverse$ denote the set of all AFs over $\arguniverse$.
\end{definition}
For two arguments $\arga,\argb\in\allargs$ of an AF $\argnetdef$ we say $\arga$ attacks $\argb$ if $(\arga,\argb)\in\allattacks$. A set $E\subseteq A$ attacks a set $E'\subseteq A$ iff there are $a\in E$ and $b\in E'$ with $(a,b)\in R$. If $E$ or $E'$ are singletons, we omit brackets, \eg, we say that $a\in A$ attacks a set $E$ if $\{a\}$ attacks $E$. We define
\begin{align*}
    \plusE=\{\arga\in\allargs\mid\setE\text{ attacks }\arga\} \quad \minusE=\{\arga\in\allargs\mid\arga\text{ attacks }\setE\}
\end{align*}
Again, for singletons we omit the brackets, \eg $a^-_\argnet=\{a\}^-_\argnet$. We say a set of arguments $U$ is unattacked if $U^-_{\argnet}\subseteq{U}$ \cite{BaroniG07}. 

The \emph{restriction} $\restrict[\argnet]{\setE}$ of an AF $\argnetdef$ to a set of arguments $\setE\subseteq\allargs$ is defined as
$\restrict[\argnet]{\setE}=(\setE,\allattacks\cap(\setE\times\setE))$. The
 \emph{reduct} $F^E$ \cite{BaumannBU2020} of an AF $\argnet$ wrt.\ a set of arguments $\setE\subseteq\argnet$ is $\reduct[\argnet]{\setE}=\restrict[\argnet]{\allargs\setminus(\setE\cup\plusE)}$.

\begin{example}
\label{ex:reduct}
For the AF $F$ below the reduct $F^{\{a\}}$ of $F$ wrt. to the singleton $\{a\}$ is the restriction of $F$ to the arguments $c,d$ and the attack $(c,d)$.
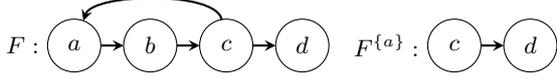
\begin{center}
    		\begin{tikzpicture}[>=stealth]
    			\node[args,label=left:$F:$] (arg1) at (0,0) {$a$};
    			\node[args] (arg2) at (1,0) {$b$};
    			\node[args] (arg3) at (2,0) {$c$};
    			\node[args] (arg4) at (3,0) {$d$};
    			\node[args,label=left:$F^{\{a\}}:$] (arg5) at (5,0) {$c$};
    			\node[args] (arg6) at (6,0) {$d$};
    			    	    			
    			\path [->,thick]	
    			(arg1) edge (arg2)
    			(arg2) edge (arg3)
    			(arg3) edge[bend right, out=280,in=250, looseness=0.55] (arg1)
    			(arg3) edge (arg4)
    			(arg5) edge (arg6)
    		
    			;
    		\end{tikzpicture}
    	\end{center}
    
\end{example}

An \emph{argumentation semantics} is a mapping $\asemantics:\argnetuniverse\to 2^{2^{\arguniverse}}$. A set $E\in\asemantics(\argnet)$ is also called a \emph{$\asemantics$-extension}. We write $\sigma_1=\sigma_2$, if $\sigma_1(\argnet)=\sigma_2(\argnet)$ for all $\argnet\in\argnetuniverse$, and $\sigma_1\subseteq\sigma_2$, if  $\sigma_1(F)\subseteq \sigma_2(F)$ for all $F\in\argnetuniverse$.

Let $\argnetdef$ be an AF. A set $\setE\subseteq\allargs$ is \emph{conflict-free} if $\setE\cap{\plusE}=\emptyset$.
We say a set $E$ defends an argument $a$ if $a^-_{\argnet}\subseteq\plusE$ and define the defense operator $\Gamma(E)=\{a\in\allargs\mid E\text{ defends }a\}$.
We consider the following semantics \cite{Baroni:2018a}.\footnote{Note that $\cfree$ and $\cladm$ are usually not regarded as (full) ``semantics'', but we treat them here to simplify language.}
{\small\begin{align*}
    \cfree(\argnet) & =\{\setE\subseteq\allargs\mid\setE\text{ is conflict-free}\}\\[-0.15cm]
    &\quad\;\;\text{\footnotesize(conflict-free semantics)}\\
    \naiv(\argnet)&= \{\setE\in\cfree(\argnet)\mid\neg \exists\setE'\in\cfree(\argnet):\setE\subsetneq \setE'\}\\[-0.15cm]
    &\quad\;\;\text{\footnotesize(naive semantics)}\\
    \cladm(\argnet)&=\{E\in\cfree(\argnet)\mid E\subseteq\Gamma(E)\}\\[-0.15cm]
    &\quad\;\;\text{\footnotesize(admissible semantics)}\\
     \clcom(\argnet)&=\{E\in\cfree(\argnet)\mid E=\Gamma(E)\}\\[-0.15cm]
    &\quad\;\;\text{\footnotesize(complete semantics)}\\
    \clpref(\argnet)&=\{E\in\clcom(\argnet)\mid \neg\exists E'\in\clcom(\argnet):E\subsetneq E'\}\\[-0.15cm]
    &\quad\;\;\text{\footnotesize(preferred semantics)}\\
    \clgrd(\argnet)&=\{E\in\clcom(\argnet)\mid \forall E'\in\clcom(\argnet):E\subseteq E'\}\\[-0.15cm]
    &\quad\;\;\text{\footnotesize(grounded semantics)}\\
        \clid(\argnet) & =\{\text{max}_{\subseteq}\{E\in\clcom(\argnet)|\: \forall E'\in\clpref(\argnet)\!:\!E\subseteq E'\}\}\\[-0.15cm]
    &\quad\;\;\text{\footnotesize(ideal semantics)}
    \end{align*}
    \begin{align*}
    \clstb(\argnet) &=\{E\in\cfree(\argnet)\mid \allargs\setminus{E}=\plusE\}\\[-0.15cm]
    &\quad\;\;\text{\footnotesize(stable semantics)}\\    
    \clsst(\argnet)&=\{E\in\clpref(\argnet)\mid \neg\exists E'\in\clpref(\argnet):\\
        & \qquad\qquad\qquad\qquad(E\cup\plusE)\subsetneq (E'\cup (E')^+_F)\}\\[-0.15cm]
    &\quad\;\;\text{\footnotesize(semi-stable semantics)}
\end{align*}}

\section{Instantiations of vacuous reduct semantics}
\label{sec:sem}
We recall the general definition of a vacuous reduct semantics from \cite{BluemelThimmKR23}. It combines two semantics $\sigma$ and $\tau$ by using one as the \emph{base condition}, which is that only $\sigma$-extensions are accepted, and one for the \emph{vacuity condition}, stating that accepted extensions have no nonempty $\tau$-extension in their reduct.
\begin{definition}
    \label{Def:vacuous-semantics}
    Let $\sigma,\tau$ be two extension-based argumentation semantics. The \emph{$\tau$-vacuous reduct semantics} $\vac_\sigma(\tau)$ to the base of $\sigma$ is defined as $$\vac_\sigma(\tau)\,(\argnet)=\{\setE\in\sigma(\argnet)\;|\;\tau(\reduct[\argnet]{\setE})\subseteq\{\emptyset\}\}$$ for all AFs $F$.
\end{definition}
The original motivation \cite{ThimmAAAI23} for vacuous reduct semantics was to define a semantics that addresses the problem of \emph{irrelevant attackers}. In \cite{ThimmAAAI23} only very few instantiations of $\vac_\sigma(\tau)$ were considered, in particular focusing on $\sigma=\cf$ and $\tau=\adm$, leading to the \emph{undisputed semantics} $\ud=\vac_\cf(\adm)$.
    \begin{example}
    \label{ex:3-cycle-attacks-1}
    In the AF $F$ from Example \ref{ex:reduct} the argument $d$ is not acceptable because it is attacked by $c$ and has no direct counterattack. However, $c$ is part of an isolated odd length cycle and therefore never accepted itself. One can argue that in this situation $d$ should be acceptable and indeed we have $\{d\}\in\ud(F)=\vac_\cf(\adm)(F)$ as $\{d\}$ is conflict-free and $F^{\{d\}}$ consists only of the cycle containing arguments $a,b,c$, for which $\emptyset$ is the only admissible set.
    \end{example}
The vacuous reduct scheme provides us with a method to refine a given semantics $\sigma$ by requesting the reduct of any extension to satisfy the vacuity condition $\tau$. When an AF $F$ satisfies $\tau(F)\subseteq\{\emptyset\}$, we also say $\tau$-vacuity is given. Vacuous reduct semantics can be seen as generalisations of the stable semantics - instead of having no arguments at all in the reduct (stable semantics) they demand no arguments accepted by a certain semantics are present. The result is a considerable degree of freedom regarding the behaviour of a vacuous reduct semantics (see Section \ref{sec:princ}). Some additional instantiations of $\vac_\sigma(\tau)$ were considered in \cite{BluemelThimmKR23}. The aim of the rest of this section is to complete the picture, by considering all different instantiations with the semantics from Section~\ref{sec:prelim}. Our findings can be divided into genuinely new semantics or correspondence to an existing or one of the new semantics. To sum up, we introduce 17 new semantics and show 31 correspondency results.
\begin{table}
\begin{center}\small
\begin{tabular}{c|c|c|c|c|c|c}
$\sigma\backslash^{\tau}$ & $\adm$ & $\gr$ & $\id$ & $\clstb$ & $\sst$ & $\cfree$ \\\hline

$\adm$ & $\pr$ \hfill $(\dagger)$ & $\co$ \hfill $(\ddagger)$ & \cellcolor{black!10} $\adm^*_1$ &\cellcolor{black!10} $\adm^*_2$ &\cellcolor{black!10} \pr  &\cellcolor{black!10} $\adm^*_3$ \\
\co &\cellcolor{black!10} $\pr$ &\cellcolor{black!10} \co &\cellcolor{black!10} $\adm^*_1$ &\cellcolor{black!10} $\co^*_1$ &\cellcolor{black!10}\pr &\cellcolor{black!10} $\adm^*_3$ \\
\pr  & $\pr$\hfill $(\ddagger)$&\cellcolor{black!10} \pr &\cellcolor{black!10} \pr &\cellcolor{black!10} \pr & \cellcolor{black!10}\pr & \cellcolor{black!10} $\adm^*_3$\\
\gr  &\cellcolor{black!10} $\gr^*_1$ &\cellcolor{black!10} \gr &\cellcolor{black!10} $\gr^*_2$ & \cellcolor{black!10}$\gr^*_3$ &\cellcolor{black!10} $\gr^*_1$ & \cellcolor{black!10} $\gr^*_4$\\
\id &\cellcolor{black!10} $\id^*_1$ & \cellcolor{black!10}\id &\cellcolor{black!10} \id &\cellcolor{black!10} $\id_2^*$ &\cellcolor{black!10} $\id^*_1$ & \cellcolor{black!10}$\id^*_3$ \\
\clstb & \cellcolor{black!10}\clstb & \cellcolor{black!10}\clstb &\cellcolor{black!10} \clstb &\cellcolor{black!10} \clstb &\cellcolor{black!10} \clstb &\cellcolor{black!10} \clstb \\
\sst  & \cellcolor{black!10}\sst &\cellcolor{black!10} \sst &\cellcolor{black!10} \sst &\cellcolor{black!10} \sst &\cellcolor{black!10} \sst &\cellcolor{black!10} $\sst^*_1$\\
\cfree & $\ud$\hfill $(\dagger)$  &$\co_{ub}$ \hfill $(\ddagger)$& \cellcolor{black!10}$\cf^*_1$  &\cellcolor{black!10} $\cf^*_2$ & \cellcolor{black!10}\ud & $\clstb_{cog}$ \hfill $(\ddagger)$ \\
\na  &\cellcolor{black!10} $\na^*_1$ & \cellcolor{black!10}$\na^*_2$ & \cellcolor{black!10}$\na^*_3$ &\cellcolor{black!10} $\na^*_4$ & \cellcolor{black!10}$\na^*_1$ &\cellcolor{black!10} $\clstb_{cog}$\\
\end{tabular}
\end{center}
\caption{Characterisations of instantiations of vacuous reduct semantics; results from \protect\cite{ThimmAAAI23} are marked with $\dagger$, results from \protect\cite{BluemelThimmKR23} are marked with $\ddagger$; all new results are highlighted with a gray background and new semantics are marked with $^*$.}
\label{tab:semantics}
\end{table}
Table~\ref{tab:semantics} gives an overview on our results. Each cell contains the semantics resulting from combining the corresponding row semantics as the base condition with the column semantics as the vacuity condition. Previously discussed results are marked accordingly and semantics that do not coincide with previously known semantics are marked with $^*$. We omit columns for $\na$, $\co$ and $\pr$ as the column for $\na$ coincides with the column for $\cf$ and the columns for $\co$ and $\pr$ coincide with the column for $\adm$. That observation rests on the following general result from \cite{BluemelThimmKR23}:
\begin{restatable}{proposition}{PROPcredulousvacuityequi}
\label{prop:credulous-vacuity-equi}
    Let $\tau,\tau'$ and $\sigma$ be argumentation semantics.
    If $\bigcup\limits_{\setE\in\tau(\argnet)}E=\bigcup\limits_{\setE\in\tau'(\argnet)}E$ for all $\argnet\in\argnetuniverse$, then for all $\argnet\in\argnetuniverse$ it holds that $\vac_{\sigma}(\tau)(\argnet)=\vac_\sigma(\tau')(\argnet)$.
\end{restatable}
The above result gives us directly the following observation.
\begin{corollary}\label{cor:simple1}
    Let $\sigma$ be an argumentation semantics. Then $\vac_\sigma(\cfree)=\vac_\sigma(\naiv)$ and $\vac_\sigma(\cladm)=\vac_\sigma(\clcom)=\vac_\sigma(\clpref)$.
\end{corollary}

The semi-stable semantics has the same effect as admissibility as a vacuity condition, although Proposition \ref{prop:credulous-vacuity-equi} cannot be applied directly as credulous acceptance under semi-stable semantics is not the same as under admissibility.
\begin{example}
    Consider the following AF $F$. 
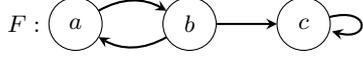
    \begin{center}
		\begin{tikzpicture}[>=stealth]
			\node[args,label=left:$F:$] (arg1) at (0,0) {$a$};
			\node[args] (arg2) at (1.5,0) {$b$};
			\node[args] (arg3) at (3,0) {$c$};
			\path [->,thick]	
			(arg1) edge[bend left] (arg2)
			(arg2) edge[bend left] (arg1)
			(arg3) edge[loop right] (arg3)
			(arg2) edge (arg3)

			;
		\end{tikzpicture}
	\end{center} Both $a$ and $b$ are credulously accepted under the classic admissible semantics but $\{b\}$ is the only semi-stable extension.
\end{example}
Nonetheless, semi-stable extensions are those admissible extensions, which minimize the argument set of the reduct, \ie $\allargs\setminus(E\cup E^+_{\argnet})$, wrt. set inclusion under all admissible extensions. So we can observe that the empty set is the only admissible extension, if and only if the empty set is also the only semi-stable extension. 
Semi-stable semantics behaves as admissibility when used in the vacuity condition, despite the fact that credulous reasoning with semi-stable semantics differs, in general, from credulous reasoning with admissibility.
\begin{restatable}{proposition}{PROPsstvacuity}
\label{prop:sst-vacuity}
Let $\sigma$ be an argumentation semantics, then $\vac_{\sigma}(\clsst)=\vac_\sigma(\cladm)$.
\end{restatable}

In the remainder of this section, we discuss the results from Table~\ref{tab:semantics} in more detail by considering each row separately.

\subsection{Admissiblity as base semantics}

In previous work, the following vacuous reduct semantics were already discussed:
\begin{restatable}{proposition}{}
\label{prop:adm-sigma}~
\begin{itemize}
    \item $\vac_{\cladm}(\sigma)=\clpref$ for $\sigma\in\{\cladm,\clcom,\clpref\}$ \cite{ThimmAAAI23,BluemelThimmKR23}
    \item $\vac_{\cladm}(\clgrd)=\clcom$ \cite{BluemelThimmKR23}
\end{itemize}
\end{restatable}

Since semi-stable vacuity is given, whenever admissible vacuity is given, we can add $\vac_{\cladm}(\clsst)$ to our list of known vacuous reduct semantics.

\begin{restatable}{corollary}{PROPadmsst}
\label{prop:adm-sst}
$\vac_{\cladm}(\clsst)=\vac_{\cladm}(\cladm)=\clpref$.
\end{restatable}

Although it is not an existing semantics, we can give a lower and upper bound for $\adm^*_1=\vac_{\cladm}(\clid)$ and make the following observations. 

\begin{restatable}{proposition}{PROPadmidobservations}
\label{prop:admidobs}~
\begin{enumerate}
\item $\clpref\subseteq\adm^*_1\subseteq\clcom$.
\item $\clid\subseteq\adm^*_1$.
\item There exist AF $\argnetdef$ such that $$\adm^*_1(\argnet)\neq\{E\in\cladm(\argnet)\mid\clid(\argnet)\subseteq E\}.$$
\end{enumerate}
\end{restatable}
We can thus think of the $\adm^*_1$ semantics as the semantics demanding that everything that can be included, without having to exclude something else, has to be included. 
The semantics works in a similar way as complete semantics. Given an admissible extension $E\supseteq\clid(F)$, $\adm^*_1$ closes $E$ under the intersection of all preferred extensions containing $E$, \ie the largest admissible extension $E'\supseteq{E}$ which is contained in all preferred supersets of $E$ is the smallest $\adm^*_1$-extension containing $E$. Following this analogy, the ideal extension is to the $\adm^*_1$-extensions what the grounded semantics is to the complete extensions, with preferred extensions being the maximal complete as well as maximal $\adm^*_1$-extensions. 

Next, let us discuss $\adm^*_2=\vac_{\cladm}(\clstb)$. To the best of our knowledge, this semantics has not been proposed yet. One can think of it as an admissible solution for cases where no stable extension exists, since there always exists at least one extension for $\vac_{\cladm}(\clstb)$.

\begin{restatable}{proposition}{PROPstbvacexistence}
\label{prop:stb-vacuity-existence}
 $\adm^*_2(\argnet)\neq\emptyset$ for all $\argnet\in\argnetuniverse$. 
\end{restatable}

A characterization of $\adm^*_2$ can be given as follows.
\begin{restatable}{proposition}{PROPadmstbchara}
\label{prop:adm-stb-chara}
Let $\argnetdef$ be an AF. Then $\adm^*_2(\argnet)=\clstb(\argnet)\cup\{E\in\cladm(\argnet)\mid\forall E'\in\clstb(\argnet):E\not\subseteq E'\}$.
\end{restatable}

In particular, it follows that every admissible extension is in $\adm^*_2$, if no stable extensions exist. On the other hand no proper subsets of stable extensions are in $\vac_{\cladm}(\clstb)$.

\begin{restatable}{corollary}{CORstbemptyvac}
\label{prop:stb-empty-vacuity}
    Let $\argnetdef$ be an AF. If $\clstb(\argnet)=\emptyset$, then $\adm^*_2(\argnet)=\cladm(\argnet)$.
\end{restatable}

As for $\adm^*_3=\vac_{\cladm}(\cfree)$, in \cite{BluemelThimmKR23} it has been mentioned that it is a proper subset of the semantics $\vac_{\cfree}(\cfree)$ (see Subsection \ref{subsec:cf}).

\subsection{Complete semantics as base semantics}
Complete semantics behaves the same way as the admissible semantics in most cases. 
First, $\vac_{\cladm}(\cladm)$, which is the preferred semantics, coincides with $\vac_{\clcom}(\cladm)$. This is due to preferred extensions always being complete. 
The same is true for $\vac_{\cladm}(\clgrd)$. Here, the vacuity condition, $\clgrd(\streduct)=\cempty$ implies there are no unattacked arguments in $\streduct$ for any extension $E\in\vac_{\cladm}(\clgrd)$, so any such $E$ is complete (Proposition \ref{prop:adm-sigma}) and thus satisfies the base condition given by complete semantics.

\begin{restatable}{proposition}{PROPcomadmcomgrd}
\label{prop:com-adm-and-com-grd}~
\begin{enumerate}
\item $\vac_{\clcom}(\sigma)=\clpref$ for any $\sigma\in\{\cladm,\clcom,\clpref,\clsst\}$.
\item $\vac_{\clcom}(\clgrd)=\clcom$
\end{enumerate}
\end{restatable}

Furthermore, since every $E\in\adm^*_1=\vac_{\cladm}(\clid)$ is a complete extension, restricting the base to complete semantics makes no difference.

\begin{restatable}{corollary}{CORcomidisadmid}
\label{cor:com-id-is-adm-id}
$\vac_{\clcom}(\clid)=\vac_{\cladm}(\clid)=\adm^*_1$.
\end{restatable}

We can also adopt the characterization of $\vac_{\cladm}(\clstb)$ for $\vac_{\clcom}(\clstb)=\clcom^*_1$, which is a new semantics.
\begin{restatable}{proposition}{PROPcomstbchara}
\label{prop:com-stb-chara}
Let $\argnetdef$ be an AF. Then $\vac_{\clcom}(\clstb)(\argnet)=\clstb(\argnet)\cup\{E\in\clcom(\argnet)\mid\forall E'\in\clstb(\argnet):E\not\subseteq E'\}$.
\end{restatable}

\begin{restatable}{corollary}{CORadmcomstb}
\label{cor:adm-com-stb}
 $\vac_{\clcom}(\clstb)\subseteq\vac_{\cladm}(\clstb)$.
\end{restatable}

Lastly, for conflict-free vacuity conditions the complete semantics as base condition yields the same reduct semantics as the admissible semantics.

\begin{restatable}{proposition}{PROPcomcfisadmcf}
Let $\sigma\in\{\cfree,\naiv\}$. Then $\vac_{\clcom}(\sigma)=\adm^*_3$.
\end{restatable}

\subsection{Preferred semantics as base semantics}

Applying admissibility-based vacuity conditions to the preferred semantics as the base condition does not yield any proper refinements, \ie every preferred extension satisfies the vacuity condition for any semantics $\sigma$ which is a subset of the admissible semantics. This is due to the fact that $\cladm(\streduct)=\cempty$ always holds for preferred extensions as shown by \cite{BaumannModularization020}.

\begin{restatable}{proposition}{PROPprfinvariant}
\label{prop:prf-invariant}
 $\vac_{\clpref}(\sigma)=\clpref$ for any $\sigma\in\{\cladm,\clcom,\clpref,\clstb,\clsst,\clgrd,\clid\}$.
\end{restatable}

Furthermore, using conflict-free vacuity conditions with preferred semantics as base condition yields the same vacuous reduct semantics as with admissible semantics as base condition, which demonstrates what a strong restriction the conflict-free vacuity condition is.

\begin{restatable}{proposition}{PROPprfcfchara}
\label{prop:prf-cf-is-admcf}
Let $\sigma\in\{\cfree,\naiv\}$. Then $\vac_{\clpref}(\sigma)=\adm^*_3$.
\end{restatable}

\subsection{Grounded semantics as base semantics}

Since the grounded semantics is a single-status semantics, refining it with a vacuity condition can only produce two outcomes: either the new semantics produces the grounded extension or no extension at all. This section is a list of simple criteria, in which cases the grounded extension is retained.

\begin{restatable}{proposition}{PROPgrdthree}
$\vac_\clgrd(\clgrd)=\clgrd$.
\end{restatable}

\begin{restatable}{proposition}{PROPgrdexistsifprf}
Let $\argnetdef$ be an AF, $\sigma\in\{\cladm,\clcom,\clpref,\clsst\}$. $\vac_\clgrd(\sigma)(\argnet)=\clgrd(\argnet)$ iff $\clgrd(\argnet)=\clpref(\argnet)$ 
\end{restatable}

\begin{restatable}{proposition}{PROPgrdstbcrit}
Let $\argnetdef$ be an AF, $G\in\clgrd(\argnet)$ the grounded extension. Then $\vac_{\clgrd}(\clstb)(\argnet)=\clgrd(\argnet)$ iff there is no $E\in\clstb(\argnet)$, such that $G\subsetneq E$.
\end{restatable}

\begin{restatable}{proposition}{PROPgrdid}
\label{prop:grd-id}
Let $\argnetdef$ be an AF. $\vac_{\clgrd}(\clid)(\argnet)=\clgrd(\argnet)$ iff $\clgrd(\argnet)=\clid(\argnet)$.
\end{restatable}

Regarding conflict-free vacuity, we have to check whether the grounded extension is in $\vac_\cfree(\cfree)(F)$. 

\begin{restatable}{proposition}{PROPgrdcfcrit}
Let $\argnetdef$ be an AF, $\sigma\in\{\cfree,\naiv\}$. Then $\vac_{\clgrd}(\sigma)(\argnet)=\clgrd(\argnet)$ iff $\clgrd(\argnet)\subseteq\vac_\cfree(\cfree)(\argnet)$.
\end{restatable}

\subsection{Ideal semantics as base semantics}

Just like the grounded semantics, the ideal semantics is a single-status semantics, so we are only concerned with its (non-)existence under the different vacuity conditions. In two cases it is guaranteed to exist.

\begin{restatable}{proposition}{PROPidgrdidid}
\begin{enumerate}
\item $\vac_{\clid}(\clgrd)=\clid$
\item $\vac_{\clid}(\clid)=\clid$
\end{enumerate}
\end{restatable}

For the other admissibility-based vacuity conditions, the cardinality of the set of preferred (resp. stable) extensions is the deciding factor.

\begin{restatable}{proposition}{PROPidadmcrit}
\label{prop:id-cardinal}
Let $\argnetdef$ be an AF.
\begin{enumerate}
\item $\vac_{\clid}(\sigma)(\argnet)=\clid(\argnet)$ for $\sigma\in\{\cladm,\clcom,\clpref,\clsst\}$ iff $|\clpref(\argnet)|=1$.
\item $\vac_{\clid}(\clstb)(\argnet)=\clid(\argnet)$ iff $|\clstb(\argnet)|\leq1$.
\end{enumerate}
\end{restatable}

Building on Propositions \ref{prop:prf-cf-is-admcf} and \ref{prop:id-cardinal}, the following can be said about the existence under conflict-free vacuity.

\begin{restatable}{proposition}{PROPidcfcrit}
Let $\argnetdef$ be an AF, $\sigma\in\{\cfree,\naiv\}$. Then  $\vac_{\clid}(\sigma)(\argnet)=\clid(\argnet)$ iff $|\clpref(\argnet)|=1$ and $\clpref(\argnet)\subseteq\vac_{\cladm}(\cfree)(\argnet)$.
\end{restatable}

\subsection{Stable and semi-stable semantics as base semantics}

Using stable semantics as the base condition reproduces stable semantics due to the reduct of a stable extension always being empty.

\begin{restatable}{proposition}{PROPstbisstb}
\label{prop:stb-base}
    Let $\argnetdef$ be an AF. For any argumentation semantics $\tau$ it holds that $\vac_{\clstb}(\tau)(\argnet)=\clstb(\argnet)$.
\end{restatable}

Semi-stable extensions are always preferred. For this reason admissibility-based vacuity conditions have no effect.

\begin{restatable}{proposition}{PROPsstadmissst}
$\vac_{\clsst}(\sigma)=\clsst$ for $\sigma\in\{\cladm,\clcom,\clgrd,\clpref,\clid,\clsst\}$.
\end{restatable}

Since the semi-stable semantics is the stable semantics on AFs where stable extensions exist, adding stable semantics as a vacuity condition has no effect, too.
\begin{restatable}{corollary}{CORsststb}
    $\vac_{\clsst}(\clstb)=\clsst$.
\end{restatable}

The only proper refinement is created by applying conflict-free vacuity. Note that there exist AFs for which $\vac_{\clsst}(\cfree)$ is a proper subset of $\vac_{\cladm}(\cfree)$.

Of course, whenever a stable extension exists, we have $\vac_{\clsst}(\cfree)(\argnet)=\clstb(\argnet)$. Apart from that, since the semi-stable extensions are defined to minimize the number of arguments in their reduct, we also get the following.
\begin{restatable}{proposition}{PROPsstcflevels}
Let $\argnetdef$ be an AF. Then for every $E\in\vac_{\cladm}(\cfree)(\argnet)$ there exists an $E'\in\vac_{\clsst}(\cfree)(\argnet)$ such that $\reduct[\argnet]{E'}$ is a restriction of $\streduct$.
\end{restatable}

So a $\cfree$-vacuous semi-stable extension exists iff a $\cfree$-vacuous admissible extension exists. The same holds in the case of naive semantics as the vacuity condition, since conflict-free vacuity and naive vacuity coincide.

\begin{corollary}
Let $\argnetdef$ be an AF. Then $\vac_{\clsst}(\sigma)(\argnet)=\emptyset$ iff $\vac_{\cladm}(\sigma)(\argnet)=\emptyset$ for $\sigma\in\{\cfree,\naiv\}$.
\end{corollary}

\subsection{Conflict-freeness as base semantics}
\label{subsec:cf}
Vacuous reduct semantics have been introduced with weak semantics in mind, \ie semantics with extensions that have attackers in their reduct. Usually, conditions are then given under which such an attacker can be ignored. By using conflict-free semantics instead of admissible semantics as the base condition, vacuous reduct semantics can realize this through different vacuity conditions and indeed, a number of weak semantics from the literature can be represented in this way \cite{BluemelThimmKR23}. 

\begin{itemize}  
    \item $\vac_{\cfree}(\sigma)$ has been introduced as the undisputed semantics \cite{ThimmAAAI23} and is the same semantics for $\sigma\in\{\cladm,\clcom,\clpref\}$ 
    \item $\vac_{\cfree}(\sigma)$ is the cogent stable semantics\footnote{derived from \cite{BodanzaT09cogsem}, defined as $\clstb_{cog}(F)=\clstb(\restrict[F]{\allargs\setminus\{a\in\allargs\mid(a,a)\in\allattacks\}}$ in \cite{BlumelUlbricht22DefeatOp}} for $\sigma\in\{\cfree,\naiv\}$ 
    \item $\vac_{\cfree}(\clgrd)$ is the ub-complete semantics \footnote{by \cite{Dondio2021ubcomplete}, characterised as $\gcom_{ub}(F)=\{E\in\cfree(\argnet)\mid\Gamma(E)\subseteq{E}\}$ by \cite{BlumelUlbricht22DefeatOp}}
\end{itemize}
Left to discuss are stable, semi-stable and ideal semantics for the vacuity condition. First, note that $\vac_{\cfree}(\clsst)$ is the undisputed semantics.
\begin{restatable}{proposition}{PROPcfsstisud}
\label{prop:cfree-semistable}
$\vac_{\cfree}(\clsst)=\vac_{\cfree}(\cladm)$.
\end{restatable}

Using the ideal semantics as the vacuity condition yields a completely new semantics, $\cfree_1^*=\vac_\cfree(\clid)$, for which the following upper and lower bound are observed.

\begin{restatable}{proposition}{PROPcfadm}
 $\vac_{\cfree}(\cladm)\subseteq\vac_{\cfree}(\clid)\subseteq\vac_{\cfree}(\clgrd)$.
\end{restatable}

$\cf^*_2=\vac_{\cfree}(\clstb)$ is also new, but features quite some odd behaviour. For instance, in the following example a completely unrelated odd cycle leads to the argument $e$ being accepted.

\begin{example}
\label{ex:3-cycle-2-chain}
In the AF $F$ the presence of the 3-cycle of the arguments $a,b,c$ makes it impossible for a stable extension to exist. Due to this $\{e\}$ has no stable extension in its reduct and is accepted under $\vac_{\cfree}(\clstb)$ despite having an unattacked attacker.
	
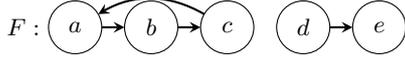
\begin{center}
		\begin{tikzpicture}[>=stealth]
			\node[args,label=left:$F:$] (arg1) at (0,0) {$a$};
			\node[args] (arg2) at (1,0) {$b$};
			\node[args] (arg3) at (2,0) {$c$};
			\node[args] (arg4) at (3,0) {$d$};
			\node[args] (arg5) at (4,0) {$e$};
			\path [->,thick]	
			(arg1) edge (arg2)
			(arg2) edge (arg3)
			(arg3) edge[bend right] (arg1)
			(arg4) edge (arg5)

			;
		\end{tikzpicture}
	\end{center}
\end{example}

Example \ref{ex:3-cycle-2-chain} also demonstrates the strange behaviour of conflict-free semantics as a vacuity condition. Consider for instance the unattacked singleton $\{d\}$, which is a preferred extension, but not an extension of $\cladm_3^*=\vac_\cladm(\cfree)$, because the odd cycle contains conflict-free sets of arguments. Together, the two observations made underline the contrast between using a very credulous semantics as the base condition, resulting in the acceptance of more arguments than one might find reasonable, and using it as a vacuity condition, which can cause excessive rejections.

\subsection{Naive semantics as base semantics}

As it was the case with preferred semantics and admissible vacuity, conflict-free vacuity is enough to ensure naivety.

\begin{restatable}{proposition}{PROPcfcfisnacf}
$\vac_{\naiv}(\sigma)=\vac_{\cfree}(\sigma)$ for $\sigma\in\{\cfree,\naiv\}$.
\end{restatable}

For the admissibility-based vacuity conditions on the other hand, naive semantics as the base condition can produce proper restrictions of the corresponding vacuous reduct semantics with conflict-free semantics as base condition.

\begin{example}
Recall Example \ref{ex:3-cycle-2-chain}. We have $\vac_{\naiv}(\sigma)=\emptyset$ for $\sigma\in\{\cladm,\clcom,\clgrd,\clpref,\clsst,\clstb\}$, since every naive extension contains a member of the odd cycle. Therefore the resulting reduct always contains an unattacked argument. In contrast, $\{d\}\in\vac_{\cfree}(\sigma)$ for all listed semantics $\sigma$, since the left-over odd cycle has no admissible extension.
\end{example}
    
\section{Principle-based Analysis}
\label{sec:princ}

We begin our analysis with two standard principles, conflict-freeness and admissibility \cite{TorreVesic17Princ}.

\begin{principle}
\label{princ:cf}
An argumentation semantics $\asemantics$ satisfies 
\textit{conflict-freeness} if $\setE\in\asemantics(\argnet)$ implies $\setE\in\cfree(\argnet)$ for every $\argnet\in\argnetuniverse$.
\end{principle}

\begin{principle}
\label{princ:adm}
An argumentation semantics $\asemantics$ satisfies 
\textit{admissibility} if $\setE\in\asemantics(\argnet)$ implies $\setE\in\cladm(\argnet)$ for every $\argnet\in\argnetuniverse$.
\end{principle}

In the case of a vacuous reduct semantics $\gvac$, the satisfaction of any of these two principles depends heavily on the choice of the base condition $\sigma$.

\begin{restatable}{proposition}{PROPcfbasedepend}
\label{prop:cf-principle}
    Let $\sigma,\tau$ be argumentation semantics. If $\sigma$ satisfies conflict-freeness (admissibility), the semantics $\vac_\sigma(\tau)$ satisfies conflict-freeness (admissibility). 
\end{restatable}

Note that conflict-freeness resp. admissibility of the semantics $\sigma$ chosen for base condition is only a sufficient, but not a necessary condition. Although the vacuity condition has no direct influence on whether an extension is conflict-free, there can be cases where the combination of base and vacuity condition results in a conflict-free semantics, while the semantics of the base condition itself is not conflict-free. We provide an example for a non-conflict-free base condition here, but the same reasoning applies to admissibility. 

\begin{example}
    Define the semantics $\sigma$ by $$\sigma(\argnet)= \left\{ \begin{array}{cc}
     \{\{a\}\}    & \text{if }\argnet=(\{a,b\},\{(a,a)\}) \\
    \cfree(\argnet)   & \text{otherwise}
    \end{array} \right.$$

Then the vacuous reduct semantics $\vac_\sigma(\cladm)$ is the undisputed semantics on all AFs which are not isomorphic to $\argnet=(\{a,b\},\{(a,a)\})$ and for this special case we get $\vac_\sigma(\cladm)(\argnet)=\emptyset$. So $\vac_\sigma(\cladm)$ produces only conflict-free extensions, while $\sigma$ does not.
\end{example}

To draw a clear line between the base condition and the vacuity condition we can make use of the principle of context-freeness, \ie the acceptance of an extension should be invariant to changes in the argument set of the surrounding AF \cite{BlumelUlbricht22DefeatOp}. 

\begin{principle}
A semantics $\sigma$ is \textit{context-free} if for any $\argnetdef,E\subseteq\allargs$ it holds that $E\in\sigma(\argnet)$ iff for all $S\subseteq\allargs,E\subseteq S:$ $E\in\sigma(\restrict[\argnet]{S})$.
\end{principle}

Examples of context-free semantics are the admissible and the conflict-free semantics. Complete semantics, on the other hand, is not context-free. Note that this principle is concerned with restrictions on the argument set, only, but makes no changes to the attacks between the considered arguments. A vacuous reduct semantics is not context-free by design, unless the vacuity condition is rendered irrelevant by the base condition, \ie is satisfied whenever the base condition is satisfied. However, when looking at the base and vacuity condition seperately, context-freeness gives us a certain degree of well-behavedness. 

\begin{restatable}{proposition}{PROPcontext}
Let $\sigma,\tau$ be argumentation semantics. If $\gvac$ is conflict-free and $\sigma$ is context-free, then $\sigma$ is conflict-free.
\end{restatable}

We now turn towards reinstatement \cite{TorreVesic17Princ}, for which two very clear sufficient conditions can be given.
\begin{principle}
\label{princ:reinstate}
An argumentation semantics $\asemantics$ satisfies \emph{reinstatement} if for every $\argnet\in\argnetuniverse$ $\setE\in\asemantics(\argnet)$ implies $\arga\in\setE$ for all $\arga$ defended by $\setE$.
\end{principle}

\begin{restatable}{proposition}{PROPreinstatement}
\label{prop:reinstatement}
Let $\sigma,\tau$ be argumentation semantics.  If
\begin{enumerate}
\item $\sigma$ satisfies reinstatement OR
\item $\sigma(\argnet)\subseteq\cfree(\argnet)$ and $a^-=\emptyset$ implies $a\in\bigcup\limits_{\setE\in\tau(\argnet)}E$ for any $a\in\allargs,\argnetdef$ an AF 
\end{enumerate}
then $\vac_\sigma(\tau)$ satisfies reinstatement.
\end{restatable}

Next, we discuss two reduct-related principles, modularization and meaningless reduct \cite{BaumannModularization020}.

\begin{principle}
\label{princ:modularization}
An argumentation semantics $\asemantics$ satisfies 
\textit{modularization} if $\setE\in\asemantics(\argnet)$ and $\setE'\in\asemantics(\reduct[\argnet]{\setE})$ implies $\setE\cup{\setE'}\in\asemantics(\argnet)$ for every $\argnet\in\argnetuniverse$.
\end{principle}

\begin{principle}
\label{princ:meaningless reduct}
An argumentation semantics $\asemantics$ satisfies 
\textit{meaningless reduct} if $\setE\in\asemantics(\argnet)$ implies $\asemantics(\reduct[\argnet]{\setE})\subseteq\{\emptyset\}$\footnote{variation of the original definition, $\subseteq$ instead of $=$} for every $\argnet\in\argnetuniverse$.
\end{principle}

A straightforward sufficient condition for both is the following:
\begin{restatable}{proposition}{PROPModulMean}
\label{prop:modularization-meanreduct}
  Let $\sigma,\tau$ be argumentation semantics. If $\gvac(\argnet)\subseteq\tau(\argnet)$ for any AF $\argnet\in\argnetuniverse$, then $\vac_\sigma(\tau)$ satisfies meaningless reduct and modularization. In particular, this holds if $\sigma(\argnet)\subseteq\tau(\argnet)$.   
\end{restatable}

With regard to modularization this is a rather trivial result. On the other hand modularization might not be all that desirable, since vacous reduct semantics aim to establish that the remaining arguments can be ignored. Therefore satisfaction of meaningless reduct can be seen as one of the more desirable principles to be satisfied by a vacuous reduct semantics. 
 A property worthy of discussion is existence of extensions \cite{TorreVesic17Princ}.
\begin{principle}
\label{princ:existence}
An argumentation semantics $\asemantics$ satisfies \emph{existence} if $\sigma(\argnet)\neq\emptyset$ for any $\argnet\in\argnetuniverse$.
\end{principle} 

First of all, a necessary condition for the non-emptiness of $\gvac(\argnet)$ is the existence of $\sigma$-extensions.

 \begin{restatable}{proposition}{PROPexistencenec}
     Let $\argnetdef$ be an AF and $\sigma,\tau$ argumentation semantics. If $\gvac(\argnet)\neq\emptyset$ then $\sigma(\argnet)\neq\emptyset$.
 \end{restatable}

Things get slightly more involved when looking for a sufficient condition. Of course, if $\tau(F)\subseteq\{\emptyset\}$ for any AF $F$ in the first place, the existence of $\sigma$-extensions suffices, but whenever $\tau$ is non-trivial we have to show at least one $\sigma$-extension for which $\tau$-vacuity of the reduct is given can be found. One way to do this is to take some $\setE\in\sigma(\argnet)$ and if $\setE\notin\gvac(\argnet)$ we simply add arguments from the reduct $\streduct$ to $\setE$ until the vacuity condition is satisfied. What we need for this is the modularization property. With $\sigma$ satisfying modularization we get:

\begin{restatable}{lemma}{LEMexistence}
\label{lem:existence}
    Let $\argnetdef$ be an AF, $\sigma$ an argumentation semantics. If $\sigma(\argnet)\neq\emptyset$ and $\sigma$ satisfies modularization, then $\vac_\sigma(\sigma)(\argnet)\neq\emptyset$.
\end{restatable}

From the lemma it follows that if we put a simple restriction on $\tau$ we can guarantee $\gvac(\argnet)\neq\emptyset$. Namely, we require $\tau$ to be stricter than $\sigma$, \ie $\tau(\argnet)\subseteq\sigma(\argnet)$ for any AF. This ensures we can take advantage of modularization to construct an extension satisfying the vacuity condition.

\begin{restatable}{proposition}{PROPexistsuff}
     Let $\sigma,\tau$ be argumentation semantics. If $\sigma$ satisfies existence and modularization and $\tau(\argnet)\subseteq\sigma(\argnet)$ for each AF $\argnet\in\argnetuniverse$, then $\vac_\sigma(\tau)(\argnet)\neq\emptyset$ for each $\argnet\in\argnetuniverse$, \ie $\vac_\sigma(\tau)$ satisfies existence.
\end{restatable}

We now consider the \emph{single-status} principle \cite{TorreVesic17Princ}. 

\begin{principle}
\label{princ:singlestatus}
An argumentation semantics $\asemantics$ satisfies \emph{single-status} if $|\asemantics(\argnet)|=1$ for every $\argnet\in\argnetuniverse$.
\end{principle}

Of course, if $\sigma$ is a single-status semantics to begin with and $\tau(\argnet)\subseteq\{\emptyset\}$ never accepts any arguments, we maintain single-status. However, if $\sigma$ is not single-status we run into two problems. First, the reduct can be the same AF for two different extensions $E,E'\in\sigma(\argnet)$. So the semantics $\sigma$ would have to explicitly disallow extensions which produce the same reduct, \eg by satisfying I-maximality (see Principle \ref{princ:Imax}) and some additional requirements. But that is only the beginning. It would also be necessary for $\tau$ to satisfy the vacuity condition on only one of the resulting reducts, which is a very unnatural property. To conclude, while vacuous reduct semantics are a refinement of their base semantics, they are not suited for choosing a single best solution.

Using modularization, a sufficient criterion for I-maximality can be given \cite{TorreVesic17Princ}. 
\begin{principle}
\label{princ:Imax}
An argumentation semantics $\asemantics$ satisfies \emph{I-maximality} if for every $\argnet\in\argnetuniverse$ and any $\setE,\setD\in\asemantics(\argnet)$ it holds that $\setE\subseteq\setD$ implies $\setE=\setD$.
\end{principle}

\begin{restatable}{proposition}{PROPimaxsuff}
    Let $\sigma,\tau$ be argumentation semantics. If 
    \begin{enumerate}
        \item $\sigma$ satisfies I-maximality OR
        \item $\sigma(\argnet)\subseteq\tau(\argnet)$ and $E,E\cup{E'}\in\sigma(\argnet)$ implies $E'\in\sigma(\streduct)$ for each AF $\argnet\in\argnetuniverse$
    \end{enumerate}
    then $\vac_\sigma(\tau)$ satisfies I-maximality.
\end{restatable}

Regarding a necessary criterion, we run into the same problem as with admissibility, without going into detail here.

As for Abstention, the idea of this principle is that extensions exist which do not contain certain disputable arguments \cite{TorreVesic17Princ}. Opposed to this, the point of reduct semantics is that the reduct of an extension does not contain any relevant arguments anymore, so we note that this principle might not be desirable for a reduct semantics.

\begin{principle}
\label{princ:abstent}
An argumentation semantics $\asemantics$ satisfies \emph{abstention} if for every $\argnet\in\argnetuniverse$ whenever there exist $E,E'\in\asemantics(F)$ such that $a\in{E}\wedge{a}\in{E'}^+$ then there exists an $E''\in\asemantics(F)$ with $a\notin(E''\cup E''^+)$.
\end{principle}

 Unlike the other properties, we do not provide a sufficient criterion here, as the requirement for $\tau$ is not quite clear for the general case. Note however, that it is a minimum requirement $\sigma$ satisfies abstention in this case.

\begin{restatable}{proposition}{PROPabstention}
\label{prop:abstention-necc}
Let $\sigma,\tau$ be argumentation semantics. If $\gvac$ satisfies abstention, then $\sigma$ satisfies abstention, too.
\end{restatable}

\emph{Directionality} demands that the extensions of a semantics $\sigma$ on an unattacked subset $U$ of an AF are exactly the restrictions of the $\sigma$-extensions of the whole AF on $U$ \cite{TorreVesic17Princ}. 

\begin{principle}
\label{princ:direct}
An argumentation semantics $\asemantics$ satisfies \emph{directionality} if for every $\argnet\in\argnetuniverse$ and for any unattacked $U\subseteq\allargs$, it holds that $\asemantics(\restrict[\argnet]{U})=\{E\cap{U}\mid{E}\in\asemantics(\argnet)\}$.
\end{principle}
Vacous reduct semantics do not satisfy directionality in general since checking the vacuity condition for a reduct wrt. to the AF as a whole usually produces a different result than for a reduct wrt. the restriction to $U$. So even if both $\tau$ and $\sigma$ satisfy directionality, this might not be true for $\gvac$. An example of this is the semantics $\vac_{\cladm}(\cfree)$.

\begin{example}
    Let $\argnet=(\{a,b\},\{(a,b),(a,a)\})$ be an AF. We have $\vac_{\cladm}(\cfree)(\argnet)=\emptyset$, but if we restrict the AF to the empty AF $\restrict[\argnet]{\emptyset}$, which is unattacked, then $\emptyset\in\vac_{\cladm}(\cfree)(\restrict[\argnet]{\emptyset})$.
\end{example}

For the satisfaction of directionality we therefore need some additional conditions. The main issue is that there might not be a superset that satisfies the vacuity condition in the whole framework. This can be solved in the same way as the existence of extensions in general, by requiring $\sigma$ to satisfy modularization and $\tau$ to be stricter than $\sigma$. We end up with the following sufficient condition:

\begin{restatable}{proposition}{PROPdirectional}
\label{prop:directionality}
 Let $\sigma,\tau$ be argumentation semantics. If
    \begin{enumerate}
        \item $\sigma$ and $\tau$ both satisfy directionality AND
        \item $\sigma$ satisfies modularization AND 
        \item $\tau(\argnet)\subseteq\sigma(\argnet)$ for each AF $\argnet\in\argnetuniverse$
    \end{enumerate} 
    then $\vac_\sigma(\tau)$ satisfies directionality.
\end{restatable}

\begin{proof}
    We have to show that for any AF $\argnetdef$ and any unattacked set $U\subseteq\allargs$ it holds that $\gvac(\restrict[\argnet]{U})=\{\setE\cap{U}\mid\setE\in\gvac(\argnet)\}$.
    ($\supseteq$) Let $E\in\gvac(\argnet)$ and $U\subseteq{\allargs}$ an unattacked set. Then $E\cap{U}\in\sigma(\restrict[\argnet]{U})$ by the directionality of $\sigma$. Furthermore, $\reduct[{\restrict[\argnet]{U}}]{E\cap{U}}$ is an unattacked subset of $\reduct[\argnet]{E}$. Since $\tau$ satisfies directionality we therefore have $\tau(\reduct[{\restrict[\argnet]{U}}]{E\cap{U}})=\tau(\reduct[\argnet]{E})\cap{U}\subseteq\{\emptyset\}$, so $E\cap{U}\in\gvac(\restrict[\argnet]{U})$.
       
       ($\subseteq$) Let $E\in\gvac(\restrict[\argnet]{U})$. Then by directionality of $\sigma$ there exists a superset $E'\in\sigma(\argnet)$ of $E$ and for any $D\in\tau(\reduct[F]{E'})$ it holds that $E'\cup{D}\in\sigma(\argnet)$ due to modularization and $\tau(\reduct[\argnet]{E'})\subseteq\sigma(\reduct[\argnet]{E'})$. By repeating the application of modularization we can therefore construct an $E''\in\gvac(\argnet)$ such that $E=E''\cap{U}$, since the finiteness of $\argnet$ guarantees that the vacuity condition is satisfied for a sufficiently large $E''$. Note that, since $\tau$ satisfies directionality and $\tau(\reduct[\restrict[\argnet]{U}]{E})\subseteq\{\emptyset\}$, we have $D\cap{U}=\emptyset$ for any $D\in\tau(\reduct[F]{E'})$, so only arguments from $\allargs\setminus{U}$ are added when constructing $E''$.
\end{proof}

Last, we discuss properties concerned with ignoring certain types of arguments. We begin with self-attack neglection which requires the set of extensions to be the same when all self-attackers are deleted from the AF \cite{BlumelUlbricht22DefeatOp}.

\begin{principle}
\label{princ:neglectself}
An argumentation semantics $\asemantics$ satisfies \emph{neglection of self-attackers} if $\asemantics(\argnet)=\asemantics(\restrict[\argnet]{\allargs\setminus\{\arga\in\allargs\mid(\arga,\arga)\in\allattacks\}})$ for every $\argnet\in\argnetuniverse$.
\end{principle}

For this principle it suffices that both semantics satisfy it.

\begin{restatable}{proposition}{PROPselfneglect}
    Let $\sigma,\tau$ be argumentation semantics. If both $\sigma$ and $\tau$  satisfy neglection of self-attackers, then $\vac_\sigma(\tau)$ satisfies neglection of self-attackers. 
\end{restatable}

Matters are not that simple for the separation property \cite{BlumelUlbricht22DefeatOp}. 
\begin{principle}
\label{princ:separation}
An argumentation semantics $\asemantics$ satisfies the \emph{separation property} if for every $\argnet\in\argnetuniverse$ and any unattacked set of arguments $U\subseteq\allargs$ with $\asemantics(\restrict[\argnet]{U})\subseteq\{\emptyset\}$\footnote{variation of the original definition, $\subseteq$ instead of $=$} it holds that $\asemantics(\argnet)=\asemantics(\restrict[\argnet]{\allargs\setminus{U}})$.
\end{principle}

Since vacuous reduct semantics are a refinement, the non-existence of $\gvac$-extensions on some unattacked set $U$ does not give us non-existence of $\sigma$-extensions on $U$, so even if $\sigma$ and $\tau$ satisfy the separation property their vacuous reduct semantics might not. 
For instance, the conflict-free semantics does satisfy the \emph{separation property} while the semantics $\vac_{\cfree}(\cfree)$ does not. 

\begin{example}
Consider $F$ from Example \ref{ex:reduct}. Let $U=\{a,b,c\}$, an odd cycle, be the unattacked set, then $\vac_{\cfree}(\cfree)(\restrict[\argnet]{U})=\emptyset$, since none of its conflict-free subsets has an empty reduct. The odd cycle transfers this property to all other conflict-free sets of $\argnet$, \eg $\{a,d\}$ has $c$ in its reduct. So $\vac_{\cfree}(\cfree)(\argnet)=\emptyset$. If, on the other hand, the odd cycle $U$ is not present, the grounded extension $G$ of the remaining AF $\restrict[F]{{A\setminus U}}=(\{d\},\emptyset$ has an empty reduct and is therefore a $\vac_{\cfree}(\cfree)$-extension on the restricted AF $\restrict[F]{{A\setminus U}}$.     
\end{example}

The non-satisfaction of the separation property in the general case is an indicator of the close relationship vacuous reduct semantics have with stable semantics. If $\tau$ is a semantics that satisfies \emph{directionality} and no set in $U$ satisfies the vacuity condition, the non-satisfaction of the vacuity condition persists for the AF as a whole, since for any extension $E$ the part of the reduct involving $U$ is not changed by adding arguments outside of $U$ to $E$.

We conclude this section with a principle-based analysis of undisputed semantics.

\begin{restatable}{proposition}{THundisputedpba}
The undisputed semantics $\vac_{\cfree}(\cladm)$ satisfies \emph{existence, conflict-freeness, reinstatement} and \emph{directionality}. It does not satisfy \emph{admissibility, context-freeness, modularization, meaningless reduct, single-status, I-maximality, abstention, neglection of self-attackers} or the \emph{separation property}.
\end{restatable}

\section{Discussion}
\label{sec:disc}

Since its introduction \cite{BaroniG07}, the principle-based approach has become the base for comparing argumentation semantics against each other. In particular, the works \cite{BaumannModularization020,DauphinRT20qualifiedsem,Dondio2021ubcomplete} compare their semantics based on weak admissibility against Dung's admissibility-based semantics using the principle-based approach.
For better comparison with these recent works 
we chose both general principles from \cite{TorreVesic17Princ} as well as principles closer related to the behaviour of weak semantics \cite{BaumannModularization020,BlumelUlbricht22DefeatOp} for our analysis. The list of principles discussed here is not exhaustive, though, one may consult \cite{Cramer19SCF2,DauphinRT20qualifiedsem} for more specific principles related to weak argumentation semantics.

As the number of new semantics yielded by applying the vacuous reduct scheme is quite large, we provide general criteria for principle satisfaction based on the two underlying semantics instead of discussing each semantics separately and only conduct an exemplary analysis for undisputed semantics. In a similar way, \cite{BlumelUlbricht22DefeatOp} focus on results concerning principle satisfaction by refute-based semantics in general. While not discussed in the original work \cite{BaroniGG05SCC}, another class of semantics with well-studied properties  are SCC-recursive semantics \cite{DauphinRT20qualifiedsem}, due to their common construction scheme of building extensions subsequently by walking through the strongly connected components of the AF. Building on these investigations, the guaranteed satisfaction of certain principles will become a considerable benefit of using general construction schemes for semantics, adding to other merits like easing recognizability and comparability of new proposals. A promising future work direction is to link the existing schemes, allowing for an efficient treatment of closely related proposals like \cite{jelia23}.

\section{Conclusion}
\label{sec:conc}
 We gave an overview on vacuous reduct semantics obtained by combining different admissibility-based and conflict-free semantics, including stable, semi-stable, ideal and naive semantics. We presented a principle-based analysis of vacuous reduct semantics in general and provided criteria for the inheritance of principle satisfaction by a vacuous reduct semantics from its base and vacuity condition for established, as well as recently introduced principles in the context of weak argumentation semantics. In particular, we discussed criteria for the existence of extensions for a given vacuous reduct semantics. We also conducted a principle-based analysis in the special case of undisputed semantics. Future work includes applying these results to the similar class of semantics proposed by \cite{jelia23} and introducing vacuous reduct semantics to structured argumentation formalisms such as, e.\,g., \emph{assumption-based argumentation} \cite{Cyras:2018}.

\section*{Ethical Statement}

There are no ethical issues.



\bibliography{usedrefsonly}

~\newpage


\appendix
   
\section{Omitted Proofs of Section \ref{sec:sem}}

\PROPsstvacuity*
\begin{proof}
($\supseteq$) Let $E\in\vac_\sigma(\cladm)$ and assume to the contrary that some $E'\in\clsst(\streduct),E'\neq\emptyset$ exists. Then $E'\in\cladm(\streduct)$, but by the vacuity condition $\cladm(\streduct)=\{\emptyset\}$. Contradiction.

($\subseteq$) Let $E\in\vac_\sigma(\clsst)$ and assume to the contrary that some $E'\in\cladm(\streduct),E'\neq\emptyset$ exists. Then there exists a nonempty superset $E''\in\clpref(\streduct)$ of $E'$ and for that a semi-stable extension $E^*\in\clsst(\streduct)$, 
such that $\reduct[\argnet]{E^{E''}}\subseteq\reduct[\argnet]{E^{E^*}}$. Since $E''\neq\emptyset$ it follows $E^*\neq\emptyset$, but by the vacuity condition $\clsst(\streduct)=\{\emptyset\}$. Contradiction.
\end{proof}

\PROPadmsst*
\begin{proof}
Follows from Prop. \ref{prop:sst-vacuity}.
\end{proof}



\PROPadmidobservations*
\begin{proof}
(1) Let $E\in\clpref(\argnet)$, then $E\in\vac_{\cladm}(\cladm)(\argnet)$, so $\cladm(\streduct)=\cempty$. But then $\clid(\streduct)=\cempty$, because the ideal extension is admissible, so $E\in\vac_{\cladm}(\clid)(\argnet)$.

For the other relation suppose to the contrary that some  $\setE\in\vac_{\cladm}(\clid)(\argnet)\setminus\clcom(\argnet)$ exists. Such an $\setE$ is admissible and at least one argument $a\in\Gamma(E)\setminus\setE$ defended by $\setE$ is not in $\setE$ (otherwise $\setE$ would be complete). Since $a^-\subseteq\plusE$, this argument is unattacked in $\streduct$, so $\clgrd(\streduct)\not\subseteq\cempty$. Since the ideal extension of the reduct must contain its grounded extension, it follows $\clid(\streduct)\not\subseteq\cempty$, so $E\notin\vac_{\cladm}(\clid)(\argnet)$. Contradiction.

 (2) Suppose to the contrary $E\in\clid(\argnet)$ and $\clid(\reduct[\argnet]{E}\not\subseteq\{\emptyset\}$, let $E'$ be the nonempty ideal extension of $\reduct[\argnet]{E}$. Then by modularization of the classic admissible semantics \cite{BaumannModularization020}, $E\cup{E'}$ is admissible. Let $P$ be a preferred extension of $\argnet$, then $E$ is a subset of $P$, so $P$ can be written as $E\cup{P'}$ for some $P'\in\clpref(\reduct[\argnet]{E})$, since $P$ surely is conflict-free and can only be preferred if $P'$ is preferred in $\reduct[\argnet]{E}$ (due to modularization). Since $P'$ is preferred in $\reduct[\argnet]{E}$, it contains $E'$. So any preferred extension $P$ of $\argnet$ contains the admissible set $E\cup{E'}$. This contradicts the maximality of the ideal extension $E$.

(3) Consider the following example:
\begin{example}
		\label{ex:id-behaviour}
		Consider the following AF $F$ with $\clid(F)=\{\emptyset\}$ and $E=\{b\}\in\clcom(F)$.
		\begin{center}
			\begin{tikzpicture}[>=stealth]
				\node[] (label) at (-1,0) {$F:$};
				\node[args] (arg1) at (-0.3,0) {$a$};
				\node[args] (arg2) at (1,0) {$b$};
				\node[args] (arg3) at (2.3,0) {$c$};
				\node[args] (arg4) at (3.6,0) {$d$};
				
				\path [->,thick]
				(arg1) edge (arg2)
				       edge[bend left] (arg3)
				(arg2) edge[bend left] (arg1)
   				(arg3) edge[bend left,in=110,out=70,, looseness=0.55] (arg1)
   				       edge (arg4)
				(arg4) edge[loop right] (arg4)
				       edge[bend left] (arg3)
				;
			\end{tikzpicture}
		\end{center}
		It holds that $\clid(\streduct)=\{\{c\}\}\neq\cempty$, so $E\notin\adm^*_1(F)$.
	\end{example}
\end{proof}

\PROPstbvacexistence*
\begin{proof}
It always holds that $\cladm(\argnet)\neq\emptyset$. There are two cases: If some $E\in\clstb(\argnet)$ exists, then $E\in\cladm(\argnet)$ and $\reduct[\argnet]{E}=(\emptyset,\emptyset)$, so $E\in\vac_{\cladm}(\clstb)(\argnet)$. If $\clstb(\argnet)=\emptyset$, then $\clstb(\reduct[\argnet]{E})=\emptyset$ for any $E\in\cladm(\argnet)$. To see this, suppose the contrary, let $E\in\cladm(\argnet)$ and $E'\in\clstb(\reduct[\argnet]{E})$. Then, since $E$ is admissible, $E\cup{E'}$ is admissible due to Modularization and stable, because $E'$ is stable in $\reduct[\argnet]{E}$ and $E$ is stable in $\argnet\!\!\downarrow_{E\cup\plusE}$. So $E\cup{E'}\in\clstb(\argnet)$. Contradiction.
\end{proof}

\PROPadmstbchara*
\begin{proof}
($\subseteq$) The base condition gives us $\vac_{\cladm}(\clstb)(\argnet)\subseteq\cladm(\argnet)$. It is left to show there are no admissible $E$ with proper stable supersets in $\vac_{\cladm}(\clstb)$. Suppose to the contrary some $E\in\vac_{\cladm}(\clstb)(\argnet)$ exists with $E\subsetneq{E'}$ for some $E'\in\clstb(\argnet)$. For such an $E$ it holds that $E'\setminus{E}\in\clstb(\reduct[\argnet]{E})$(Suppose the contrary, then some argument $a\in\reduct[\argnet]{E}\setminus(E'\setminus{E}\cup{E'^+_\argnet}\setminus{\plusE})$ exists. But then $E'$ is not stable in $\argnet$, since $E$ does not attack arguments in $\reduct[\argnet]{E}$. Contradiction.). But then, since $E$ is a proper subset of $E'$, we have $E'\setminus E\neq\emptyset$ and therefore $\clstb(\streduct)\neq\cempty$. So $E\notin\vac_{\cladm}(\clstb)(\argnet)$. Contradiction.

($\supseteq$) Let $E\in\clstb(\argnet)$, then $E\in\cladm(\argnet)$ and $\reduct[\argnet]{E}=(\emptyset,\emptyset)$ has no nonempty stable extensions, so $E\in\vac_{\cladm}(\clstb)(\argnet)$. For the other case suppose to the contrary $E\in\{E\in\cladm(\argnet)\mid\forall E'\in\clstb(\argnet):E\not\subseteq E'\}$ and some non-empty $E'\in\clstb(\reduct[\argnet]{E})$ exists. Then, since $E$ is at least admissible, $E\cup{E'}$ is admissible in $\argnet$ due to modularization and stable there, because $E'$ is stable in $\reduct[\argnet]{E}$ and $E$ is stable in $\argnet\!\!\downarrow_{E\cup\plusE}$. So $E\cup{E'}\in\clstb(\argnet)$ is a proper stable superset of $E$. Contradiction.
\end{proof}

\CORstbemptyvac*
\begin{proof}
  Follows directly from Proposition \ref{prop:adm-stb-chara}.
\end{proof}

\PROPcomadmcomgrd*
\begin{proof}
\begin{enumerate}
\item ($\subseteq$) Let $E\in\vac_{\clcom}(\sigma)(\argnet)$, then $E\in\vac_{\cladm}(\sigma)$ since the vacuity conditions coincide and every complete extension is admissible. By Propositions \ref{prop:adm-sigma} and \ref{prop:adm-sst} follows that $E\in\clpref(\argnet)$.

($\supseteq$) Let $E\in\clpref(\argnet)$, then $E\in\clcom(\argnet)$ and $E\in\cap\vac_{\cladm}(\sigma)(\argnet)$ by \cite{ThimmAAAI23}, so $E\in\vac_{\clcom}(\sigma)$.

\item ($\subseteq$) Analogously to the first case, if $E\in\vac_{\clcom}(\clgrd)(\argnet)$ then $E\in\vac_{\cladm}(\clgrd)(\argnet)=\clcom(\argnet)$, as shown in \cite{BluemelThimmKR23}. For ($\supseteq$) let $E\in\clcom(\argnet)$. Then $E$ contains all arguments, which it defends, so $\streduct$ does not contain any unattacked arguments, therefore $\clgrd(\streduct)=\{\emptyset\}$ and $E\in\vac_{\clcom}(\clgrd)$.
\end{enumerate}
\end{proof}

\CORcomidisadmid*
\begin{proof}
Follows directly from Proposition \ref{prop:admidobs}.
\end{proof}

\PROPcomstbchara*
\begin{proof}
Analogously to the proof of Proposition \ref{prop:adm-stb-chara}, the considerations for the vacuity condition apply in the same way.
\end{proof}

\CORadmcomstb*
\begin{proof}
Follows from Propositions \ref{prop:adm-stb-chara}, \ref{prop:com-stb-chara} and the fact that $\clcom(\argnet)\subseteq\cladm(\argnet)$.
\end{proof}

\PROPcomcfisadmcf*
\begin{proof}
($\subseteq$) follows from $\clcom(\argnet)\subseteq\cladm(\argnet)$ for any $\argnetdef$.

($\supseteq$) Suppose to the contrary $E\in\vac_{\cladm}(\sigma)$ and $E\notin\clcom(\argnet)$. Then there is at least one argument $a$ in the reduct $\streduct$, that is defended by $E$ and therefore unattacked in $\streduct$. But then $\{a\}$ is a non-empty conflict-free extension of $\streduct$, which has a naive superset, so $E\notin\vac_{\cladm}(\sigma)$. Contradiction.
\end{proof}

\PROPprfinvariant*
\begin{proof}
($\subseteq$) per Definition \ref{Def:vacuous-semantics}, ($\supseteq$) follows from Proposition 3.2 in \cite{BaumannModularization020}, since $\sigma(\argnet)\subseteq\cladm(\argnet)$ for all considered $\sigma$.
\end{proof}

\PROPprfcfchara*
\begin{proof}
($\subseteq$) follows from $\clpref(\argnet)\subseteq\cladm(\argnet)$ for any $\argnetdef$.

($\supseteq$) Suppose to the contrary $E\in\vac_{\cladm}(\sigma)$ and $E\notin\clpref(\argnet)$. Since $E$ is admissible, it has a preferred superset and modularization allows us to represent this superset as $E\cup{E'}$, where $E'\in\cladm(\streduct)$. Thus, there exists a nonempty admissible extension of $\streduct$. But then $E'$ is also a non-empty conflictfree extension of $\streduct$, which has a naiv superset, so $E\notin\vac_{\cladm}(\sigma)$. Contradiction.
\end{proof}

\PROPgrdthree*
\begin{proof}
($\subseteq$) By Definition \ref{Def:vacuous-semantics}. ($\supseteq$) The grounded extension of an AF, let it be $E$, is a complete extension, therefore  $\streduct$ contains no unattacked arguments, so $\clgrd(\streduct)=\cempty$.
\end{proof}

\PROPgrdexistsifprf*
\begin{proof}
($\Rightarrow$) Let $E$ be the grounded extension of $\argnet$. If $E\in\vac_{\clgrd}(\sigma)$ for any of these $\sigma$, it follows that $\cladm(\streduct)=\cempty$. By Proposition 3.2 in \cite{BaumannModularization020} therefore $E\in\clpref(F)$. The grounded extension is a subset of all preferred extensions. Therefore, if it is preferred, it has to be the only preferred extension, as it cannot have any proper admissible supersets. 

($\Leftarrow$) Let $E$ be the grounded extension of $\argnet$. If $E\in\clpref(\argnet)$ then by Proposition 3.2 in \cite{BaumannModularization020} it follows that $\cempty=\cladm(\streduct)\supseteq\sigma(\streduct)$ for all $\sigma$ considered.
\end{proof}

\PROPgrdstbcrit*
\begin{proof}
($\Rightarrow$) Suppose there is an $E\in\clstb(\argnet)$ with $G$ as a proper subset. Then $E\setminus{G}\in\clstb(\reduct[\argnet]{G})$ is not empty. Contradiction with the vacuity condition.

($\Leftarrow$) Suppose $G\notin\vac_{\clgrd}(\clstb)(\argnet)$ and there is no $E\in\clstb(\argnet)$, such that $G\subsetneq E$. Then there exists a stable, non-empty $S\in\clstb(\reduct[\argnet]{G})$. But then by Modularization $G\cup{S}\in\clstb(\argnet)$ and therefore a proper superset of $G$ is stable. Contradiction.
\end{proof}

\PROPgrdid*
\begin{proof}
($\Rightarrow$) Let $E$ be the grounded extension of $\argnet$, which is also ideal, and suppose to the contrary $\clid(\streduct)=\{I\}\neq\cempty$. Then, since $E$ is the ideal extension, due to modularization every preferred extension can be written as $E\cup{P}$ with $P$ being preferred in $\streduct$. Therefore, $E\cup{I}$ is a subset of each preferred extension $E\cup{P}$ of $\argnet$. Furthermore, since complete semantics satisfies modularization and both the grounded and the ideal semantics are complete, it follows $E\cup{I}\in\clcom(\argnet)$. But then $E$ was not maximal and therefore not ideal to begin with. Contradiction.
($\Leftarrow$) The grounded extension $E$ is a subset of the ideal extension $I$. Assume to the contrary $E\subsetneq{I}$, then due to modularization $I\setminus{E}$ is the non-empty ideal extension of $\streduct$, so $E\notin\vac_{\clgrd}(\clid)(\argnet)$. Contradiction.  
\end{proof}

\PROPgrdcfcrit*
\begin{proof}
Let $G$ be the grounded extension.
($\Rightarrow$) If $G\in\vac_{\clgrd}(\sigma)(\argnet)$, then $G$ satisfies the conflict-free vacuity condition and since the grounded extension is admissible we have $G\in\vac_{\cladm}(\cfree)(\argnet)$.
($\Leftarrow$) If $G$ in $\vac_{\cladm}(\cfree)$, then by Corollary \ref{cor:simple1} it satisfies the vacuity condition for both $\sigma$ and since it is the grounded extension, we have $G\in\vac_{\clgrd}(\sigma)(\argnet)$.
\end{proof}

\PROPidgrdidid*
\begin{proof}
\begin{enumerate}
\item Follows from the fact that the ideal extension is complete (see the proof of Prop. \ref{prop:com-adm-and-com-grd}, item 2) 
\item Analogously to the proof of Proposition \ref{prop:grd-id}.
\end{enumerate}
\end{proof}

\PROPidadmcrit*
\begin{proof}
Let $E$ be the ideal extension of $\argnet$.
\begin{enumerate}
\item ($\Rightarrow$) Suppose there are distinct $P,P'\in\clpref(\argnet)$, then $P\setminus{E}$ is a non-empty admissible extension of $\streduct$. Contradiction to the vacuity condition.

($\Leftarrow$) When there is only one preferred extension, that extension is the ideal extension and as a preferred extension also satisfies the vacuity condition due to Proposition 3.2 in \cite{BaumannModularization020}.

\item ($\Rightarrow$) Analogously to (1, $\Rightarrow$).

($\Leftarrow$) Every stable extension is preferred. Therefore, if there is one stable extension, Proposition 3.2 in \cite{BaumannModularization020} applies. If $|\clstb(\argnet)|=0$, suppose to the contrary some nonempty $S\in\clstb(\streduct)$ existed, then $E\cup{S}\in\clstb(\argnet)$ due to modularization. Contradiction.\qedhere
\end{enumerate}
\end{proof}

\PROPidcfcrit*
\begin{proof}
($\Rightarrow$) Since $\cfree(\streduct)=\cempty$ implies $\cladm(\streduct)=\cempty$ this follows from the proof of Proposition \ref{prop:id-cardinal}.

($\Leftarrow$) Follows from the fact, that in this case $\clid(\argnet)=\clpref(\argnet)$ and the single extension satisfies the vacuity condition. 
\end{proof}

\PROPstbisstb*
\begin{proof}
    By Definition~\ref{Def:vacuous-semantics} $(\subseteq)$ holds, it is left to show $(\supseteq)$. Let $E\in\clstb(\argnet)$ then $E$ attacks all arguments not in $E$ by the definition of stable semantics, so $E\cup\plusE=\allargs$ and $\reduct[\argnet]{E}=(\emptyset,\emptyset)$. Therefore $\tau(\reduct[\argnet]{E})\subseteq\{\emptyset\}$. Thus, $E$ satisfies the vacuity condition, so $E\in\vac_{\clstb}(\tau)(\argnet)$.
\end{proof}

\PROPsstadmissst*
\begin{proof}
Follows directly from Prop. \ref{prop:prf-invariant}, since $\clsst(\argnet)\subseteq\clpref(\argnet)$ for any AF $\argnetdef$.
\end{proof}

\CORsststb*
\begin{proof}
    Follows directly from Propositions~\ref{prop:stb-base} and \ref{prop:stb-empty-vacuity}.
\end{proof}

\PROPsstcflevels*
\begin{proof}
let $E\in\vac_{\cladm}(\cfree)(\argnet)$. From the definition of semi-stable semantics it follows some $E'\in\clsst(\argnet)$ exists, such that $\reduct[\argnet]{E'}$ is a subframework of $\streduct$. But then also $\cfree(\reduct[\argnet]{E'})\subseteq\cfree(\streduct)=\cempty$, so $E'\in\vac_{\clsst}(\cfree)(\argnet)$. 
\end{proof}

\PROPcfsstisud*
\begin{proof}
Follows from Prop. \ref{prop:sst-vacuity}.
\end{proof}

\PROPcfadm*
\begin{proof}
Let $E\in\vac_{\cfree}(\cladm)(\argnet)$, then $\cladm(\streduct)=\{\emptyset\}=\clid(\streduct)$, so $E\in\vac_{\cfree}(\clid)(\argnet)$. Now let $E\in\vac_{\cfree}(\clid)(\argnet)$. Then $\clid(\streduct)=\{\emptyset\}$ so the ideal extension is $\emptyset$. Since the ideal extension is a complete extension and the grounded extension is a subset of every complete extension, it follows $\clgrd(\streduct)=\emptyset$ and $E\in\vac_{\cfree}(\clgrd)(\argnet)$. 
\end{proof}
\PROPcfcfisnacf*
\begin{proof}
($\subseteq$) follows from the conflict-freeness of naive extensions. For ($\supseteq$) suppose to the contrary an $E\in\vac_{\cfree}(\cfree)(\argnet)$ existed, which is not naive. Then some argument $a\in\allargs\setminus E$ must exist, such that $E\cup\{a\}$ is conflict-free. We know $a\notin\plusE$, so $a\in\streduct$ and not a self-attacker. But then $\{a\}\in\cfree(\streduct)$. Contradiction. 
\end{proof}

\section{Omitted Proofs of Section \ref{sec:princ}}

\PROPcfbasedepend*
\begin{proof}
    Follows directly from Definition~\ref{Def:vacuous-semantics}, every $E\in\vac_\sigma(\tau)(\argnet)$ is a $\sigma$-extension and thus conflict-free, analogously for admissibility. 
\end{proof}

\PROPcontext*
\begin{proof}
Suppose to the contrary an $E\in\sigma(\argnet)$ exists for some AF $\argnetdef$, which is not conflict-free. Then $E\in\sigma(\restrict[\argnet]{E})$, in which case $\tau\subseteq\{\empty\}$, because $\reduct[\restrict[\argnet]{E}]{E}$ contains no arguments. Therefore $E\in\vac_\sigma(\tau)(\restrict[\argnet]{E})$, so $\vac_\sigma(\tau)$ is not conflict-free.
\end{proof}

\PROPreinstatement*
\begin{proof}
\begin{enumerate}
\item Follows from the subset-relation $\vac_\sigma(\tau)(\argnet)\subseteq\sigma(\argnet)$.
\item For any $\setE\in\cfree(\argnet)$ it holds that $\Gamma(E)\setminus{E}$ is the set of unattacked arguments in $\reduct[\argnet]{\setE}$, because the attackers of any such argument must all be in $\plusE$. So if $\tau$ accepts unattacked arguments, then $\tau(\reduct[\argnet]{\setE})\not\subseteq\{\emptyset\}$ unless there are no unattacked arguments in $\streduct$
\end{enumerate}
\end{proof}

\PROPModulMean*
\begin{proof}
    Let $E\in\gvac(\argnet)$, then, since $\gvac(\streduct)\subseteq\tau(\streduct)$ it follows from the vacuity condition $\tau(\streduct)\subseteq\{\emptyset\}$ that $\gvac(\streduct)\subseteq\{\emptyset\}$, so $\gvac$ satisfies meaningless reduct and due to that modularization as well. The special case follows from $\gvac(\argnet)\subseteq\sigma(\argnet)$.
\end{proof}

\PROPexistencenec*
 \begin{proof}
    Follows from $\gvac(\argnet)\subseteq\sigma(\argnet)$. 
 \end{proof}
 
 \LEMexistence*
\begin{proof}
    Let $\setE\in\sigma(\argnet)$. If $\sigma(\streduct)\subseteq\{\emptyset\}$, we are done. If not, some nonempty $E'\in\sigma(\streduct)$ exists. By modularization it then follows $\setE\cup\setE'\in\sigma(\argnet)$. Since $\argnet$ has only finitely many arguments, we can extend $\setE$ in this way until the reduct no longer contains any $\sigma$-extensions, worst case, until it is empty.
\end{proof}

\PROPexistsuff*
\begin{proof}
    Follows from Lemma \ref{lem:existence} and the fact that $\sigma(\streduct)\subseteq\{\emptyset\}$ implies $\tau(\streduct)\subseteq\{\emptyset\}$ due to the third condition $\tau(\argnet)\subseteq\sigma(\argnet)$.
\end{proof}

\PROPimaxsuff*
\begin{proof}
    \begin{enumerate}
        \item Follows from $\gvac(\argnet)\subseteq\sigma(\argnet)$ for any AF.
        \item Assume to the contrary that $E,E\cup{E'}\in\sigma(\argnet)$ implies $E'\in\sigma(\streduct)$ for each AF $\argnet\in\argnetuniverse$ holds and $\sigma(\argnet)\subseteq\tau(\argnet)$ but for some AF $\argnetdef$ there exist $\setE,\setE'\in\gvac(\argnet)$ such that $E\subsetneq{E^*}$. Then $E^*=E\cup(E^*\setminus{E})$, so $E^*\setminus{E}\in\sigma(\streduct)$ is not empty. Now since $\sigma(\streduct)\subseteq\tau(\streduct)$ it follows that $\tau(\streduct)\not\subseteq\{\emptyset\}$, so $E\notin\gvac(\argnet)$. Contradiction.
    \end{enumerate}
\end{proof}

\PROPabstention*
\begin{proof}
Follows from $\gvac(\argnet)\subseteq\sigma(\argnet)$.
\end{proof}

\PROPselfneglect*
\begin{proof}
    Since $\sigma$ satisfies neglection of self-attackers we have $E\in\sigma(\argnet)$ iff $E\in\sigma(\restrict[\argnet]{\allargs\setminus\{a\in\allargs\mid(a,a)\in\attacks\}})$. In particular any such $E$ does not contain self-attackers, therefore $\restrict[\streduct]{\allargs\setminus\{a\in\allargs\mid(a,a)\in\attacks\}}=\reduct[\restrict[\argnet]{\allargs\setminus\{a\in\allargs\mid(a,a)\in\attacks\}}]{E}$. It follows that the vacuity condition is satisfied for $E$ in $\argnet$ if it is satisfied in $\restrict[\argnet]{\allargs\setminus\{a\in\allargs\mid(a,a)\in\attacks\}}$, since $\tau$ satisfies self-attack neglection. So $E\in\gvac(\argnet)$ iff $E\in\gvac(\restrict[\argnet]{\allargs\setminus\{a\in\allargs\mid(a,a)\in\attacks\}})$.
\end{proof}

\THundisputedpba*
\begin{proof}
(Conflict-freeness) The base condition is the conflict-free semantics, so by Proposition~\ref{prop:cf-principle}.

(Admissibility)
Consider $F=(\{a,b\},\{(a,a),(a,b)\})$ where $\{b\}$ is undisputed for a counterexample.

(Reinstatement) Yes, since $\sigma$ is the conflict-free semantics, by \ref{prop:reinstatement} it suffices that tau, which is the admissible semantics, always accepts unattacked arguments.

(Abstention) As a counterexample use $F=(\{a,b\},\{(a,b),(b,a)\})$. We have $\{a\},\{b\}\in\vac_{\cf}(adm)$ but no extension $E$ for which $a$ is "undecided", \ie neither in nor attacked by it.

(Single-Status) As a counterexample use again $F=(\{a,b\},\{(a,b),(b,a)\})$ with $\{a\},\{b\}\in\vac_{\cf}(adm)$.

(I-maximality) Consider $F=(\{a,b\},\{(a,a),(a,b)\})$ where both the empty set and $\{b\}$ are undisputed.

(Directionality) To be concise: Prop. \ref{prop:directionality} applies. 

($\Leftarrow$) Suppose $E\in\vac_{\cf}(\adm)(F)$ and $U\subseteq{\allargs}$ an unattacked set. Then $E\cap{U}$ is conflictfree. Furthermore, $\reduct[{\restrict[\argnet]{U}}]{E\cap{U}}$ is an unattacked subset of $\reduct[\argnet]{E}$. Since the admissible semantics satisfies directionality we therefore have $\adm(\reduct[{\restrict[\argnet]{U}}]{E\cap{U}})=\adm(\reduct[\argnet]{E})\cap{U}=\{\emptyset\}$, so $E\cap{U}$ is undisputed in $\restrict[\argnet]{U}$.
       
       ($\Rightarrow$) Let $E\in\vac_{\cf}(\adm)(\restrict[\argnet]{U})$. Then for any $E'\in\pr(\reduct[F]{E})$ it holds that $E\cup{E'}\in\vac_{\cf}(\adm)(F)$. $E\cup{E'}$ is conflict-free, since $E'\cap{U}=\emptyset$ and $E'\subseteq\reduct[F]{E}$ and the vacuity condition follows from the fact that $E'$ is preferred, due to modularization therefore $\adm(\reduct[F]{E\cup{E'}})=\{\emptyset\}$.
       
(Modularization/Meaningless Reduct) As a counterexample consider $F=(\{a,b,c\},\{(a,a),(a,b),(b,c)\})$. Then $\{c\}$ is undisputed, $\{b\}$ is undisputed in the reduct, violating meaningless reduct, and $\{b,c\}\notin\vac_{\cf}(\adm)(F)$, violating modularization.

(Self-Attack Neglection/Separation Property) Consider again $F=(\{a,b,c\},\{(a,a),(a,b),(b,c)\})$. Then $\{c\}$ is undisputed, but not undisputed in $F^\circ=(\{b,c\},\{(b,c)\})$. Since $\{a\}$ is an unattacked subset of $F$ with no nonempty undisputed extension, this also shows the separation property does not hold.
\end{proof}

\end{document}